\documentclass[pdflatex,sn-mathphys-num]{sn-jnl}

 \usepackage{booktabs}

\usepackage{multicol}
\usepackage{multirow}
\usepackage{longtable}
\usepackage{enumitem}
\usepackage{graphicx}
\usepackage{xcolor}
\usepackage{marvosym}
\usepackage{url}
\usepackage{algorithm}
\usepackage{algpseudocode}
\usepackage{amsmath}
\usepackage{lmodern}
\usepackage{hyperref}
\usepackage{adjustbox}

 \usepackage{soul}

\usepackage{tikz,collcell}
\usetikzlibrary{shapes,arrows}
\usetikzlibrary{datavisualization.formats.functions}
\usetikzlibrary{calc,trees,positioning,arrows,chains,shapes.geometric,shapes,shadows,matrix,patterns}
\usetikzlibrary{hobby,calc}
\usetikzlibrary{decorations.markings}

\begin{document}

\title[NER for the Kurdish Sorani Language: Dataset Creation
and Comparative Analysis]{Named Entity Recognition for the Kurdish Sorani Language: Dataset Creation
and Comparative Analysis}

\author[1]{\fnm{Bakhtawar} \sur{Abdalla}}\email{bakhtawar.abdalla@spu.edu.iq}

\author[2,1]{\fnm{Rebwar Mala} \sur{Nabi}}\email{rebwar.nabi@kti.edu.iq}

\author[3]{\fnm{Hassan} \sur{Eshkiki}}\email{h.g.eshkiki@swansea.ac.uk}

\author*[3]{\fnm{Fabio} \sur{Caraffini}}\email{fabio.caraffini@swansea.ac.uk}

\affil[1]{Technical College of Informatics, Sulaimani Polytechnic University, \city{Sulaimaniyah}, \state{Sulaimaniyah}, \country{Iraq}}

\affil[2]{Kurdistan Technical Institute, \city{Kurdistan region}, \state{Sulaimaniyah}, \country{Iraq}}

 \affil[3]{\orgdiv{Department of Computer Science}, \orgname{Swansea University}, \city{Swansea}, \postcode{SA1 8EN}, \country{UK}}

\abstract{This work contributes towards balancing the inclusivity and global applicability of natural language processing techniques by proposing the first `name entity recognition' dataset for Kurdish Sorani, a low-resource and under-represented language, that consists of $64,563$ annotated tokens. It also provides a tool for facilitating this task in this and many other languages and performs a thorough comparative analysis, including classic machine learning models and neural systems. The results obtained challenge established assumptions about the advantage of neural approaches within the context of NLP. Conventional methods, in particular CRF, obtain F1-scores of $0.825$, outperforming the results of BiLSTM-based models ($0.706$) significantly. These findings indicate that simpler and more computationally efficient classical frameworks can outperform neural architectures in low-resource settings.}


\maketitle

\section{Introduction}
\label{sec:intro}

Named Entity Recognition (NER) is the activity of identifying and labelling proper nouns within a text, such as the names of persons, places, organisations, and dates. The task supports a wide range of natural language processing (NLP) applications, from information extraction, machine translation, and question-answering systems \cite{Joshi2020TheSA}, to sentiment analysis \cite{Kumar2023AComprehensive} and text summarisation \cite{Wang2023SurveyingThe}, to name a few.

Over the past decade, NLP research has garnered significant attention and achieved remarkable performance levels with the English language. Numerous neural systems in the literature consistently report F1-scores exceeding 93\% in datasets such as the popular Conference on Computational Natural Language Learning (CoNLL) CoNLL03~\cite{tjong2003introduction}. Nevertheless, this does not accurately reflect the present state of the art in many other languages, which constitutes a significant limitation, especially given that some of the most impactful, responsible, and promising applications of NLP involve enabling machine translation and other assistive technologies for the benefit of all individuals. In this context, NER is crucial in NLP, yet it faces support challenges for many languages due to a lack of annotated corpora. This imbalance results in a technological disadvantage for millions of speakers.

In this light, researchers started proposing NER datasets to fill the gap in different languages. There have been significant efforts to create comprehensive NER resources on high-resource European languages. In the case of Spanish, the CoNLL-2002 shared task offered initial datasets\cite{carreras2002named}, which were later extended by a corpus of a specific domain. French NER was developed based on such resources as the ALEDA dataset~\cite{sagot2012aleda} and several news-based corpora that delivered competitive performance rates. Improvement of German NER development was significant with the NoSta-D dataset~\cite{benikova2014nosta}, and advancement with multilingual frameworks. These high-resource languages have attained F1-scores of 85-90\% in many domains in reading comprehension, creating strong baselines and allowing cross-lingual transfer learning strategies. Nevertheless, the challenges in NER remain high in the case of low-resource languages such as Kurdish Sorani since they lack the available annotated datasets, whereas high-resource languages have gained considerable progress. This shortage is not only a barrier to creating individualised NLP applications but also to making cross-linguistic comparative linguistic studies and improving cross-lingual learning strategies.

Kurdish Sorani, spoken by approximately 8 million people, is a good example of a widely spoken yet under-represented language in computational linguistics research \cite{ahmadi2020klpt}. There are multiple reasons behind this under-representation. 
\begin{itemize}
    \item First, Sorani uses a modified Arabic script without standardised orthographic rules. 
    \item Second, the language exhibits complex morphology through agglutination and cliticisation, as it forms complex words by combining roots, prefixes, and suffixes. 
    \item Third, contemporary texts frequently mix Arabic and Persian elements \cite{Mccarus1997NationalismAL}.
    \item Most importantly, no publicly available NER datasets exist for Kurdish Sorani.
\end{itemize}

The lack of NER tools has direct consequences. The Kurdish regions have a vast amount of digital content, created via news websites, social media, and government websites. However, these communities cannot access automated information extraction, machine translation, or question-answering systems. This is a constraint to their engagement in the digital world and access to information networks \cite{Joshi2020TheSA}.

Recent research has shown that under limiting data conditions, existing machine learning frameworks can achieve results similar to those achieved by deep neural networks (DNN) \cite{hedderich2020survey}. However, these findings come from languages with at least some annotated resources. Since there are no Kurdish Sorani named entity recognition (NER) datasets, such systematic comparative studies of modelling paradigms are not possible.

The current study addresses this gap by making three vital contributions. 
\begin{itemize}
    \item We present the first  `AgaCKNER' dataset for Kurdish Sorani and provide a tool for manual annotation of NER supporting this (and similar) language.
    \item We perform a comparative analysis across four established models for NLP. \item We analyse how the unique linguistic characteristics of Kurdish Sorani influence the effectiveness of each NLP model.
\end{itemize}

The remainder of this article is structured as follows. 
\begin{itemize}
\item \textbf{Section~2} reviews existing NER approaches in the literature, highlighting challenges in low-resource language processing, and presenting relevant work on Kurdish language processing.
\item \textbf{Section~3} describes the dataset creation process, including data collection, preprocessing, annotation methodology, and quality validation.
\item \textbf{Section~4} presents the four comparison models and their specific configurations for this study.
\item \textbf{Section~5} details the experimental design, evaluation metrics, data splitting strategies, and reproducibility measures.
\item \textbf{Section~6} presents comprehensive performance comparison results across models and entity types.
\item \textbf{Section~7} discusses the theoretical and practical implications of our findings for low-resource NLP and Kurdish language processing.
\item \textbf{Section~8} concludes this study, summarises key contributions and findings, and discusses limitations and future work.

\end{itemize}

\section{Background}
\label{sec:background}

During the 1990s, NER became an independent NLP task at the Message Understanding Conferences (MUC-6 and MUC-7), where several standardised NER tasks were developed, including identification and labelling of proper nouns in text \cite{grishman1996message}. In the NER task, the boundaries of entities must be localised and specific predefined categories assigned (most commonly: person (PER), location (LOC), organisation (ORG), and miscellaneous entity (MISC)). Early research relied on handwritten rules and gazetteers, producing fairly low performance on well-structured corpora. A breakthrough was the introduction of statistical techniques, especially Hidden Markov Models and Conditional Random Fields, which allowed learning patterns based on annotated data as opposed to learnt rules \cite{lafferty2001conditional}.

The CoNLL-2002 and CoNLL-2003 shared tasks provided an equivalent formalisation to NER testing across languages, defining the Beginning-Inside-Outside (BIO) tagging format, which is now widely adopted \cite{tjong2003introduction}. The B-X sequences indicate the occurrence of entity type X, and the I-X tags indicate continuation, and the O tags indicate non-entity tokens. They also served as benchmark datasets in English, German, Spanish, and Dutch, as they allowed systematic comparison of NER systems. Modern neural architectures will achieve human-level results in these datasets, using transformer-based models, they commonly achieve F1-scores higher than $93\%$ \cite{devlin2019bert}.

Advances in NER have primarily focused on high-resource languages, creating an enormous hole in terms of datasets, pre-trained embeddings, and access to available linguistic tools for many languages. There are various initiatives that reduce these limitations as outlined in the literature. The ANERcorp corpus of $150,000$ tokens helped Arabic NER build systems that scored $85\%$ on the F1-score \cite{Benajiba2009ANERsysAA}. ArmanPersoNERCorpus in Persian offers $250,000$ tokens in news and social media categories \cite{poostchi2018bilstm}. The MasakhaNER project succeeded in an African-language NER by providing datasets in ten languages, such as Yoruba, Igbo, and Swahili, with about $10,000$ tokens each \cite{Adelani2021MasakhaNERNE}. In Tibetan language processing, Yu et al. proposed a rule-based person name recognition scheme based on a case-auxiliary grammar and lexicon, with $94.02\%$ precision and $90.13\%$ recall on a newspaper corpus \cite{yu2010named}.

The use of cross-lingual transfer has been proven to be a successful strategy to solve low-resource Named Entity Recognition (NER). The multilingual systems have been proved to work reasonably well in a wide range of languages, and it only needs a little adjustment, which has been supported with the actual application to Hungarian and English datasets \cite{Szarvas2006AMN}. 

Limited datasets can be expanded by data augmentation through automatic translation and entity substitution, but the quality of the data produced strongly depends on the method of choice \cite{enghoff2018low}. Active learning aims to learn to annotate informative ones also, which can significantly decrease labelling costs by $40\%$, performance can reliably be met \cite{shen2017deep}. Semi-supervised algorithms leverage large amounts of unlabelled corpora and small labelled corpora, they are improved by language-specific features \cite{liang2020bond}.

Despite significant progress in the field of natural language processing, a relatively large number of languages in the world are not represented at all. According to the taxonomy developed by Joshi et al. \cite{Joshi2020TheSA}, more than $2,000$ languages are not represented in any labelled dataset.

Kurdish presents a very challenging case in NLP due to limited resources and high levels of variability in dialects. In this regard, both the Sorani and Kurmanji variants struggle with a severe lack of data, although they differ as much by language as by any script tradition \cite{Esmaili2013SoraniKV}. Indeed, Kurmanji is widely spoken in the northern regions and uses Latin script, while Sorani is used predominantly in the central and southern parts of Kurdistan and is written in modified Arabic script and without a standard orthography in comparison to the Kurmanji. Hence, Sorani is characterised by its unique phonological, grammatical, and syntactical characteristics where the word boundaries and the spelling of words vary irregularly \cite{Mccarus1997NationalismAL}. The language shows rich morphology in the form of agglutination, which causes high surface forms caused by individual lemmas.

For these reasons, and despite its rich literary heritage, significant online presence through the news media and state communications, and a significant number of speakers, Sorani has been historically underrepresented in digital corpora and NLP research, underscoring the importance of specialised resources to foster more effective language processing and computational analysis tools \cite{Haig2014IntroductionTS}. Sorani has been undreserved in NLP as there are no standard NER datasets available, no established baseline systems, and no systematic analysis has ever been done.

\section{Preparing the `AgaCKNER' dataset}

The proposed AgaCKNER dataset consists of Kurdish Sorani text formatted according to CoNLL format with text tokens in the first column and their corresponding tags for NER in the second. Each statement ends with a period tagged as `O' (. O) and is separated by a blank line from the next statement.

The name of the dataset is the combination of the word  `Aga', a common given name and expression in Kurdistan that translates in English to `be careful', and the acronym CKNER, which stands for Central Kurdish NER. 

This is the first dataset of this kind and serves as a key resource to standardise benchmarking, advance NER systems and allow technological progress and digital inclusion for Kurdish speakers.

\begin{figure}[ht!]
\begin{adjustbox}{max width = \textwidth}
\begin{tikzpicture}[
node distance = 2mm,
  start chain = going right,
 start/.style = {signal, draw=#1, fill=#1!30,
                 text width=40mm, minimum height=28mm, font=\large,
                 signal pointer angle=150, text centered, on chain},
  cont/.style = {start=#1, signal from=west}
                 ]

\node[start=gray]{
\bfseries Data collection  \normalfont
\begin{itemize}
  \item Over $160$ articles \\reviewed
  \item $5$ news categories covered
\end{itemize}
};

\node[cont=orange] {\bfseries
                     Preprocessing
                     \normalfont
                        \begin{itemize}
                    \item Number \& date conversion
                    \item Cleaning \& \\normalisation

                        \end{itemize}   
};

\node[cont=blue] {\bfseries\centering
                     Formatting \\\& \\annotation\\ 
                     \normalfont
\begin{itemize}
                    \item CoNLL format
                    \item Original Web~app\\for annotation

\end{itemize}   
};

\node[cont=red] {\bfseries\centering
                     Quality Check \\
};

\node[draw, rectangle, text width=30mm,text centered,on chain]{\textbf{AgaCKNER dataset}};

\end{tikzpicture}
\end{adjustbox}
    \caption{Constructing the dataset: key steps.}
    \label{fig:steps}
\end{figure}
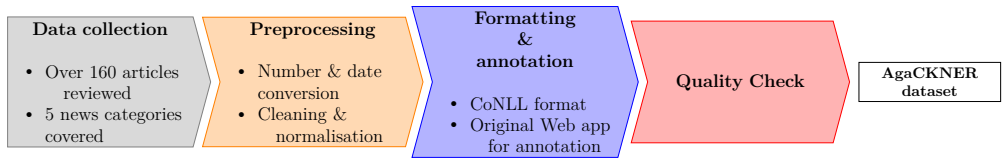

Figure \ref{fig:steps} provides an overview of the key phases for generating AgaCKNER, from raw data collection to processed and annotated data entries. Descriptions for these steps and further information on the dataset are provided in the following subsections.

\subsection{Data Collection}\label{sec:datacol}
The corpus of this pioneering dataset is gathered from the Rudaw Media Network \cite{Rudaw2025}, an independent Kurdish for-profit media establishment. This network professionally distributes written, visual, and audio content, including in Kurdish Sorani, while prioritising the maintenance of a global and impartial stance concerning geographical and ideological affiliations as a core objective.

The data collection process focussed on recent news articles published between late 2023 and early 2024 across five distinct categories: Kurdistan News, Middle East News, World News, Economic News and Sports News. The articles were selected in such a way that each category had almost 32 articles, and the final sample comprised of more than 160 articles.

To ensure the validity of the collected information, each article was meticulously selected by targeted data scraping from the web pages of the respective sources. The rationale for employing this method stems from the potential distortion or loss of the original Kurdish Sorani text, or the introduction of extraneous elements, as a result of passing through multiple encoding processes.

\subsection{Data Preprocessing}\label{sec:preprocessing}

A preprocessing phase is needed to transform the collected raw data into a workable text dataset to be structured in an NER format.

\subsubsection{Cleaning~and~Text~Normalisation}

During the initial preprocessing stage, the accumulated textual data underwent systematic cleaning and normalisation through the Kurdish Language Processing Toolkit (KLPT) \cite{ahmadi2020klpt}, a toolkit specifically designed for the processing of low-resource Kurdish dialects. This phase of preprocessing consisted of the implementation of four fundamental steps:
\begin{enumerate}
\item Removal of unwanted characters.
\item Elimination of English words and characters using regular expression pattern matching.
\item Conversion of English numerals to Kurdish Sorani numerals using our character mapping system \cite{agackner_allinone}. 
\item Word and sentence tokenisation using pertinent Kurdish Language Processing Toolkit (KLPT) modules in conjunction with Python's string manipulation functionalities.
\end{enumerate}
The aforementioned process preserved the linguistic integrity of the Kurdish Sorani text while simultaneously standardising its format. The cleaned corpus retained the semantic relationships and contextual information essential for NER, ensuring that token boundaries and numerical representations were consistent with Kurdish Sorani conventions.

The source code for the entire preprocessing pipeline is available on GitHub \cite{agackner_mendeleydataset} for replicability and to facilitate the development of Kurdish Sorani NLP tools.

\subsection{Data Format Conversion and Annotation}\label{sec:format}

After preprocessing, the curated data were formatted to comply with the CoNLL guidelines. This is a widely adopted and accepted choice in NLP settings. By presenting each token alongside its corresponding entity label in distinct rows, this format ensures compatibility with the majority of training frameworks for NER models.

We designed and implemented the web-based application `AGA NER Annotation Tool', available from GitHub \cite{agackner_allinone}, purposely for annotating textual data in CoNLL format in the Kurdish Sorani language and used it in this work.

\begin{figure}[ht!]
    \centering
    \includegraphics[width=0.55\linewidth]{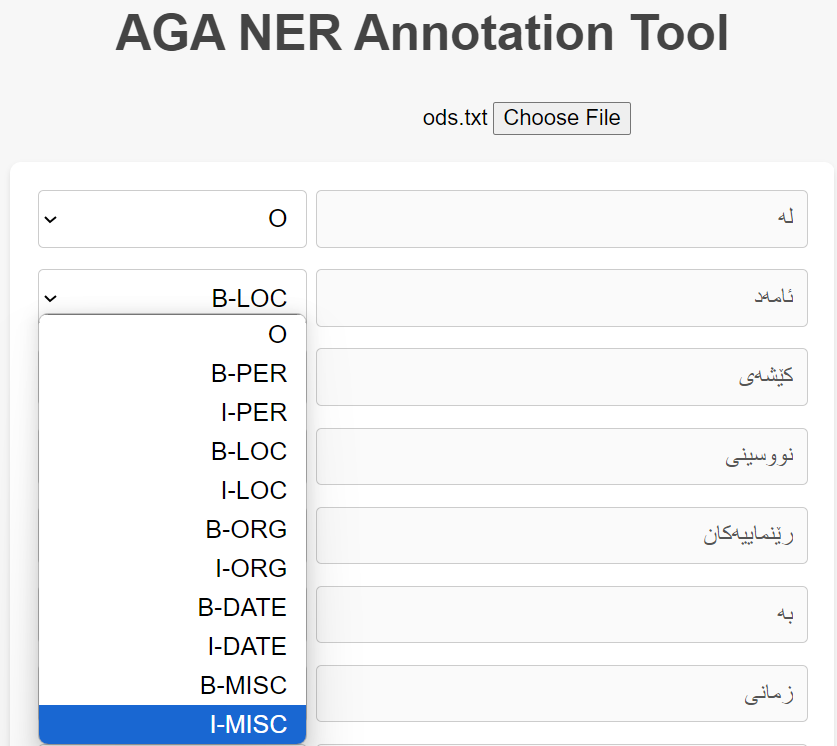}
    \caption{Interface of the AGA NER Annotation Tool.}
    \label{fig:interface}
\end{figure}

Upon initiating access to this tool, users can upload their (CoNLL) dataset to visualise it through an interface (Figure \ref{fig:interface}) showing both the text content and the designated entity type selection.

\begin{figure}[ht!]
    \centering
    \includegraphics[width=0.55\linewidth]{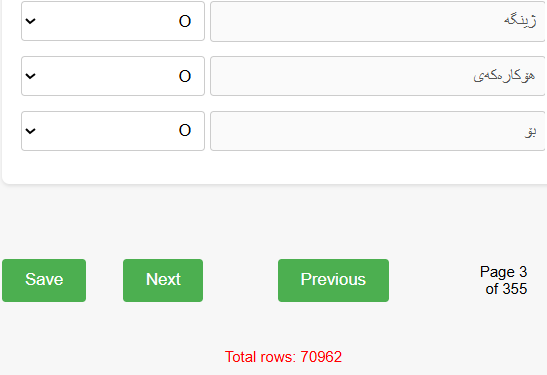}
    \caption{Navigation buttons and drop-down menus for annotation in the AGA NER Annotation Tool web-app.} 
    \label{fig:navigation}
\end{figure}

 The AGA NER Annotation Tool divides the uploaded data into subsets to show 100 rows per page. Navigation buttons (previous/next) are positioned at the bottom and at the top of each page, along with page- and row-count information (Figure \ref{fig:navigation}). 

 The keyboard shortcuts displayed in the `Key' column of Table \ref{tab:2columns} can accelerate the annotation process when used instead of the drop-down menu to assign a tag.

\begin{table}[ht!]
\large
\centering
\caption{Keyboard Shortcuts for Entity Type Selection}\label{tab:2columns}
\begin{tabular}{ccc}
\hline
\textbf{Key} & \textbf{Entity Type}& \textbf{Description} \\
\hline
1 + $\uparrow$ & O & Others \\
1 & B-PER & Beginning of Person \\
2 & I-PER & Inside of Person \\
3 & B-LOC & Beginning of Location \\
4 & I-LOC & Inside of Location \\
5 & B-ORG & Beginning of Organisation \\
6 & I-ORG & Inside of Organisation \\
7 & B-DATE & Beginning of Date \\
8 & I-DATE & Inside of Date \\
9 & B-MISC & Beginning of Miscellaneous \\
0 & I-MISC & Inside of Miscellaneous \\
\hline
\end{tabular}
\end{table}

\subsection{Validation and Inter-Annotator Agreement (IAA)}\label{sec:IAA}
As described earlier, the dataset in this study was developed through a manual annotation process. Two expert linguists, both specialising in the Kurdish language, manually assigned entity labels to the dataset. Each annotation was carefully reviewed and refined to ensure accuracy and consistency. Throughout the process, the annotators collaborated closely, resolving any disagreements through discussion and mutual consensus. This meticulous manual approach helped ensure the reliability and linguistic integrity of the dataset.
To measure the consistency between inter-annotator agreement (IAA), it was assessed using Cohen’s Kappa ($\kappa$), a widely recognised statistical metric that adjusts for chance agreement. The dataset achieved a Cohen’s Kappa score of $\kappa$ = 0.92, which falls under the “almost perfect” agreement category ($\kappa$> 0.80) by standard interpretation. This high level of agreement demonstrates the reliability and quality of the annotated dataset and highlights the effectiveness of combining automated labelling with expert validation.

\subsection{Dataset Files}

The AgaCKNER dataset comprises two files, detailed in Table~\ref{tab:dataset_files}. The primary file, \texttt{AgaCorpus.txt}, contains $2,534$ sentences and $64,563$ tokens structured in CoNLL format. Each line represents a token followed by its corresponding BIO (Beginning, Inside, Outside) label for five entity types: PERSON, LOCATION, ORGANIZATION, DATE, and MISCELLANEOUS.

\begin{table}[ht!]
\centering
\large
\caption{AgaCKNER Components.}
\label{tab:dataset_files}
\begin{tabular}{|l|p{8cm}|}
\hline
\textbf{File Name} & \textbf{Description} \\
\hline
\texttt{AgaCKNER\_Dataset.txt} & Main corpus file containing $2,534$ sentences and $64,563$ tokens in CoNLL format. Each line represents a token with its BIO label for entity types \\
\hline
\texttt{AgaCKNER\_metadata.txt} & Documentation file detailing data collection process, article sources, collection dates, and Kurdish Sorani preprocessing steps \\
\hline
\end{tabular}
\end{table}

 The use of this standardised format ensures easy processing and easy comparison with other NER datasets. The BIO tagging scheme provides strong boundaries of entities; `B-' indicates the start of an entity, `I-' indicates the continuation of the same entity, and `O' indicates that the tokens are not in any named entity.



\begin{figure}[ht!]
    \centering
    \includegraphics[keepaspectratio,width=0.35\textwidth]{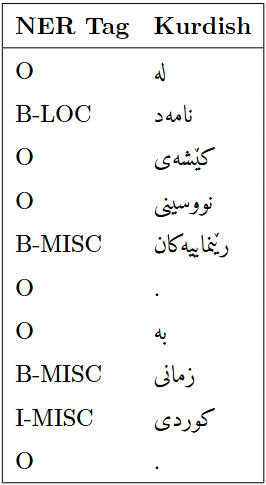}
   \caption{A fragment from \texttt{AgaCKNER\_Dataset.txt} with English translations (column order adjusted per request).}
    \label{fig:annotated}
\end{figure}

 An example of annotated text is provided in Figure \ref{fig:annotated}.


\subsection{Entity Distribution Analysis}

The AgaCKNER dataset shows a diverse distribution of named entities, as detailed in Table~\ref{tab:entity_distribution}.

\begin{table}[ht!]
\large
\centering
\caption{Entity Type Distribution}
\label{tab:entity_distribution}
\begin{tabular}{|l|r|r|}
\hline
\textbf{Entity Type} & \textbf{Occurrences} & \textbf{Percentage (\%)} \\
\hline
PERSON & 2,814 & 4.36 \\
LOCATION & 3,576 & 5.54 \\
ORGANIZATION & 4,207 & 6.52 \\
DATE & 1,532 & 2.37 \\
MISCELLANEOUS & 2,775 & 4.30 \\
OUTSIDE & 49,659 & 76.91 \\
\hline
Total & 64,563 & 100.00 \\
\hline
\end{tabular}
\end{table}

Among the $64,563$ tokens in the proposed dataset, ORGANIZATION entities exhibit the highest frequency, appearing $4,207$ times ($6.52\%$), followed by LOCATION entities, which occur $3,576$ times ($5.54\%$). PERSON entities constitute $2,814$ tokens ($4.36\%$), whereas MISCELLANEOUS entities embody $2,775$ tokens ($4.30\%$). DATE entities are observed to be the least prevalent with $1,532$ instances ($2.37\%$). A significant proportion of tokens ($76.91\%$) are classified as OUTSIDE, denoting non-entity text, which mirrors the distribution of natural language found in news articles.

The balanced distribution of entity types, despite a significant emphasis on the extensive representation of organisations and locations, renders the dataset suitable for developing robust Kurdish Sorani NER models.

\vskip 0.5cm

\section{Comparison models selection and settings}

The developed dataset forms the basis of a comparative analysis that enables derivation of baseline performance indicators of Kurdish Sorani NER, evaluates the usefulness of various modelling paradigms in the low-resource context, and provides clear guidance to practitioners in similarly under-resourced languages. The analysis is central to the general question of what approaches most appropriately address morphologically rich languages that have little annotated data and to the trade-offs in resource-limited conditions between the model complexity and performance.

There are four Kurdish Sorani NER models that we compare on the basis of their successfulness in performing sequence labelling tasks and the ability to process the limited amount of available training data. These models denote both conventional machine learning and deep learning.

\subsection{Conditional Random Fields (CRF)}

CRF is a discriminative probabilistic model particularly effective for sequence labelling tasks such as NER~\cite{lafferty2001conditional}. Unlike generative models, CRF directly models the conditional probability, $p\left(y|x\right)$ where $x$ the input token sequence is represented and $y$ the corresponding label sequence is denoted. 

For a linear-chain CRF, the conditional probability is defined as:
\begin{equation}
p\left(y|x\right) = \frac{1}{Z(x)} \exp\left(\sum_{j} \lambda_j F_j(y,x)\right)
\end{equation}

where $Z\left(x\right)$ is the normalisation factor, and $F_j\left(y,x\right) = \sum_{i=1}^{n} f_j\left(y_{i-1}, y_i, x, i\right)$ are the feature functions. Where the input is the sequence of words or tokens x, i is the position index in the sequence, and Fj (y, x) is the relationship between the corresponding current label yi and the preceding current label yi-1. Feature functions are mathematical expressions that describe possible interest patterns within the input sequence and account for dependencies among consecutive labels. Individual feature functions fj(yi-1, yi, x, i) observe the connections between the previous label yi-1, current label yi, input sequence x, and position i. The functions are able to record many linguistic patterns, including the identity of words, morphological patterns, capitalisation, prefixes, suffixes, and contextual knowledge of adjacent tokens.

\par
CRF benefits from being conditioned, so assumptions of independence are no longer required, unlike in Hidden Markov Models (HMMs). This allows for the incorporation of arbitrary, overlapping, and agglomerative observation features from both past and future contexts. For Kurdish Sorani NER, we employ both word-level and contextual features, including lowercase words, prefixes, suffixes, and neighbouring words within a $\pm2$ word window.

\subsection{Support Vector Machine (SVM)}

SVM is a supervised learning method that works particularly well in situations with many features~\cite{cortes1995support}. In identifying entities, SVM is generally employed as a binary classifier for every entity type and decides whether each word fits into a specific class.

For our Kurdish Sorani NER implementation, we use a linear kernel, which performs well for text classification problems. The model incorporates comprehensive feature extraction, including basic sub-word features (lowercase forms, prefixes, suffixes, and digits), contextual features (neighbouring words within $\pm$2 positions), and structural characteristics (word shape, length, and symbols).

SVM works very well on small to medium-sized datasets, so it is ideal for our Kurdish Sorani corpus. The system effectively deals with large numbers of language features, which are critical for NER in morphologically rich languages.

\subsection{Bidirectional Long Short Term Model (BiLSTM)}
BiLSTM is a recurrent neural network designed to analyse data sequences by processing them both backwards and forward~\cite{graves2005framewise}. As a result, the model can consider information from both the back and front parts of the input string.

In the BiLSTM architecture, two LSTM layers handle the input separately but essentially in reverse order. The states from each direction are added together to get the final representation.
\begin{equation}
h_i = [\overrightarrow{h_i}; \overleftarrow{h_i}]
\end{equation}

In our Kurdish Sorani NER task, BiLSTM builds a dense representation for each word by automatically using an embedding layer. This approach helps us bypass feature engineering and captures both semantic and syntactic kinds of relationships. The architecture manages sequences of any length by padding and masking.

\subsection{BiLSTM-CRF}
BiLSTM-CRF uses the feature learning power of BiLSTM and the structured model of CRF~\cite{huang2015bidirectional}. By using both BiLSTM and CRF, this architecture can learn context and also make sure the tags follow the right order.

The model architecture has an embedding layer, several bidirectional LSTM layers, a linear projection layer and a CRF layer. BiLSTM spots the important features from the input order, while CRF takes care of how the labels are connected and supports finding the best overall tag sequence.

This combination addresses the limitation of standalone BiLSTM models in sequence labelling tasks where label dependencies are crucial. The presence of the CRF component stops any BIO transitions that are not allowed, as an I-PER tag should not be used after B-LOC.

\subsection{Model Configurations}
The specific configurations and hyperparameters for every model used in our experiments are shown in Table~\ref{tab:model_configs}.

\begin{table}[ht!]
\centering
\caption{Model configurations and hyperparameters}
\label{tab:model_configs}
\begin{tabular}{lll}
\toprule
Model & Parameter & Value \\
\midrule
\multirow{4}{*}{CRF} & Optimization Algorithm & L-BFGS \\
 & L1 Regularisation Coefficient & 0.1 \\
 & L2 Regularisation Coefficient & 0.1 \\
 & Maximum Iterations & 200 \\
\midrule
\multirow{3}{*}{SVM} & Kernel & Linear \\
 & Feature Vectorisation & DictVectoriser \\
 & Feature Scaling & MaxAbsScaler \\
\midrule
\multirow{5}{*}{BiLSTM} & Embedding Dimension & 50 \\
 & LSTM Units & 100 \\
 & Spatial Dropout Rate & 0.1 \\
 & Recurrent Dropout & 0.1 \\
 & Sequence Length (padded) & 50 tokens \\
\midrule
\multirow{5}{*}{BiLSTM-CRF} & Embedding Dimension & 128 \\
 & BiLSTM Hidden Dimension & 256 (128 each direction) \\
 & Learning Rate & 0.0005 \\
 & Batch Size & 2 \\
 & Epochs & 10 \\
\bottomrule
\end{tabular}
\end{table}

\section{Experimental methods and setup}
\label{sec:experimental}

\subsection{Data Split Strategies}
To evaluate model robustness, we employ two data-splitting strategies.

\begin{itemize}
    \item The first strategy consists of uniformly selecting 70\% of the dataset for training and the remaining 30\% for testing.
    \item The second instead uses a 80\% for training and a 20\% for testing split.
    \item These data split ratios are common in NER research, particularly for low-resource languages ~\cite{taher2019persian,zirikly2015named,shaalan2014survey}.
\end{itemize}

Both splitting ratios help assess model performance based on training data size. The $70/30$ method provides a larger test set for thorough testing, while $80/20$ maximises the dataset size, crucial for the underresourced Kurdish Sorani language.

\subsection{Cross Validation}
To minimise bias and enhance result reliability, we implement 10-fold cross-validation for each model configuration. This method of cross-validation divides the dataset into 10 equal parts. The test set only uses one fold, while 9 other folds are used for training. We average performance metrics across all iterations to obtain robust estimates.

Cross-validation effectively manages small datasets like AgaCKNER by using available data efficiently and providing stable performance estimates. The same approach has provided good results for NER of Arabic text~\cite{benajiba2008arabic} and Persian text~\cite{poostchi2018bilstm}.

\subsection{Baseline Comparisons}
To evaluate our methods, we compare them to well-known NER approaches that have been applied in similar low-resource language contexts. Our comparison includes methods applied to Persian NER~\cite{poostchi2018bilstm} and Arabic NER~\cite{el2019arabic}, adapting their configurations to Kurdish Sorani where applicable.

\subsection{Evaluation Metrics}
We compute three established performance metrics to assess the model in terms of precision (Eq. \ref{eq:precision}), recall (Eq. \ref{eq:recall}) and F1-score (Eq. \ref{eq:f1}) as shown below:

\begin{equation}\label{eq:precision}
    \text{Precision} = \frac{\text{TP}}{\text{TP} + \text{FP}}
\end{equation}

\begin{equation}\label{eq:recall}
    \text{Recall} = \frac{\text{TP}}{\text{TP} + \text{FN}}
\end{equation}

\begin{equation}\label{eq:f1}
   \text{F1-score} = \frac{2 \times \text{Precision} \times \text{Recall}}{\text{Precision} + \text{Recall}} 
\end{equation}


where TP denotes true positives, FP denotes a false positives, and FN denotes a false negatives. These common evaluation measures are suitable for imbalanced NER datasets~\cite{li2020survey}. The F1-score is the primary metric, as it balances precision and recall, making it ideal for entity recognition tasks where both measures are important.

Table~\ref{tab:exp_config} summarises the experimental configuration.

\begin{table}[ht!]
\centering
\caption{Experimental configuration summary}
\label{tab:exp_config}
\begin{tabular}{ll}
\toprule
Configuration & Setting \\
\midrule
Data splits & 70/30, 80/20 \\
Cross-validation & 10-fold \\
Evaluation metrics & Precision, Recall, F1-score \\
Entity types & 5 (PER, LOC, ORG, DATE, MISC) \\
Annotation format & BIO tagging \\
\bottomrule
\end{tabular}
\end{table}

\subsection{Reproducibility}

For the sake of reproducibility, we provide the source code for the model implementations and the entire experimental setup through the online repository in the GitHub repository \cite{agackner_allinone}. Our original AgaCKNER dataset is made public at \cite{agackner_mendeleydataset}.


\section{Results and Analysis}\label{sec:results}

\subsection{Model Performance Comparison}

Table~\ref{tab:results} illustrates the overall performance of four NER models on AgaCKNER data using various data splitting methods and reveals the differences in the effectiveness of the traditional machine learning and deep learning approaches to Kurdish Sorani NER.

\begin{table}[ht!]
\centering
\caption{Performance Comparison of NER Models on the AgaCKNER Dataset}
\begin{tabular}{lcccc}
\toprule
Model & Data Split & Precision & Recall & F1-score \\
\midrule
CRF & 70/30 & 0.8510 & 0.7852 & \textbf{0.8168} \\
SVM & 70/30 & 0.8203 & 0.7374 & 0.7766 \\
BiLSTM & 70/30 & 0.6628 & 0.7773 & 0.7155 \\
BiLSTM-CRF & 70/30 & 0.7801 & 0.6789 & 0.7260 \\
\midrule
CRF & 80/20 & 0.8466 & 0.8049 & \textbf{0.8252} \\
SVM & 80/20 & 0.8190 & 0.7505 & 0.7833 \\
BiLSTM & 80/20 & 0.6521 & 0.7704 & 0.7063 \\
BiLSTM-CRF & 80/20 & 0.7825 & 0.7137 & 0.7465 \\
\bottomrule
\end{tabular}
\label{tab:results}
\end{table}


The F1-scores are always highest when using the CRF model, with results of 0.8168 for 70\% train, 30\% test and 0.8252 for 80\% train, 20\% test. This superior performance demonstrates the effectiveness of conditional random fields for sequence labelling in Kurdish Sorani text. The F1-score increase from 70/30 to 80/20 split (0.84 percentage points) shows that CRF profits from receiving more data for training.

Out of all algorithms, SVM achieved the second-highest F1-scores, with 0.7766 in the first split and 0.7833 in the second. The results demonstrate that the model accurately picks up the right entities, but not as often as CRF.

To validate these findings, we conducted 10-fold cross-validation experiments. Table~\ref{tab:cv_results} gives the results of the cross-validation, with information on the mean and standard deviation, which allows us to better evaluate the performance.

\begin{table}[ht!]
\centering
\caption{10-Fold Cross-Validation Results (Mean $\pm$ Standard Deviation)}
\begin{tabular}{lccc}
\toprule
Model & Precision & Recall & F1-score \\
\midrule
CRF & 0.8406 $\pm$ 0.0170 & 0.7976 $\pm$ 0.0109 & \textbf{0.8184 $\pm$ 0.0090} \\
SVM & 0.8329 $\pm$ 0.0102 & 0.7704 $\pm$ 0.0104 & 0.8004 $\pm$ 0.0088 \\
BiLSTM & 0.6699 $\pm$ 0.0258 & 0.7982 $\pm$ 0.0090 & 0.7282 $\pm$ 0.0137 \\
BiLSTM-CRF & 0.7779 $\pm$ 0.0139 & 0.7207 $\pm$ 0.0145 & 0.7481 $\pm$ 0.0117 \\
\bottomrule
\end{tabular}
\label{tab:cv_results}
\end{table}

Cross-validation results confirm the superior performance of CRF with a mean F1-score of $0.8184$ $\pm 0.0090$. The low standard deviation indicates consistent performance across different data folds. SVM demonstrates the highest stability with the lowest standard deviations across all metrics, suggesting reliable performance regardless of data distribution. We conducted paired t-tests to assess statistical significance, confirming that CRF significantly outperforms all other models ($p < 0.05$) in terms of F1-score.

\subsection{Deep Learning vs Traditional Models Analysis}
A comparison of how the methods perform uncovers some insights about deep learning and traditional machine learning in Kurdish Sorani NER. 

BiLSTM recalls a high number of relevant cases ($0.7982 \pm 0.0090$), but its precision is low ($0.6699 \pm 0.0258$). These findings show that BiLSTM can detect possible entities, but it is less accurate in labelling them, leading to more false positives.

The BiLSTM-CRF hybrid model shows improved precision compared to standalone BiLSTM ($0.7779$ vs $0.6699$) while maintaining reasonable recall. However, there is no improvement in performance over the standard CRF or SVM models. This points out that a small amount of data could prevent optimal results from deep learning, a frequent problem in low-resource language processing.

The comparison between $70/30$ and $80/20$ splits reveals that all models benefit from additional training data, with CRF showing the most substantial improvement ($0.84$ percentage points in F1-score). It should be noted that adding more data only shows little improvement for BiLSTM, possibly because the original data was already causing overfitting.

\subsection{Entity-Level Performance and Error Analysis}

\subsubsection{Detailed Entity Performance}

Table~\ref{tab:entity_detailed} shows detailed performance results for every entity type for each model on both datasets, highlighting that recognition accuracy is quite different for each entity type.

\begin{table}[ht!]
\centering
\caption{Entity-Level F1-scores by Model and Data Split}
\begin{tabular}{lcccccccc}
\toprule
\multirow{2}{*}{Entity Type} & \multicolumn{2}{c}{CRF} & \multicolumn{2}{c}{SVM} & \multicolumn{2}{c}{BiLSTM} & \multicolumn{2}{c}{BiLSTM-CRF} \\
\cmidrule(lr){2-3} \cmidrule(lr){4-5} \cmidrule(lr){6-7} \cmidrule(lr){8-9}
& 70/30 & 80/20 & 70/30 & 80/20 & 70/30 & 80/20 & 70/30 & 80/20 \\
\midrule
B-DATE & 0.77 & 0.77 & 0.71 & 0.71 & 0.63 & 0.62 & 0.67 & 0.72 \\
B-LOC & 0.84 & 0.85 & 0.81 & 0.82 & 0.76 & 0.77 & 0.77 & 0.78 \\
B-MISC & 0.71 & 0.72 & 0.69 & 0.70 & 0.62 & 0.60 & 0.62 & 0.66 \\
B-ORG & 0.83 & 0.82 & 0.79 & 0.79 & 0.73 & 0.70 & 0.78 & 0.78 \\
B-PER & 0.85 & 0.90 & 0.81 & 0.83 & 0.70 & 0.65 & 0.69 & 0.77 \\
I-DATE & 0.81 & 0.82 & 0.75 & 0.78 & 0.71 & 0.69 & 0.71 & 0.73 \\
I-LOC & 0.80 & 0.81 & 0.75 & 0.75 & 0.69 & 0.69 & 0.71 & 0.74 \\
I-MISC & 0.65 & 0.64 & 0.62 & 0.62 & 0.53 & 0.57 & 0.55 & 0.54 \\
I-ORG & 0.84 & 0.85 & 0.80 & 0.82 & 0.75 & 0.76 & 0.76 & 0.77 \\
I-PER & 0.90 & 0.92 & 0.83 & 0.82 & 0.75 & 0.69 & 0.73 & 0.76 \\
\bottomrule
\end{tabular}
\label{tab:entity_detailed}
\end{table}

\subsubsection{Entity-Specific Analysis}

\textbf{PERSON Entities:} CRF demonstrates superior performance for person recognition, achieving F1-scores of $0.85-0.90$ for B-PER and $0.90-0.92$ for I-PER tags. The high accuracy (0.91-0.94) shows that person names are likely to be well identified, with few false positive results. BiLSTM shows the lowest performance for person entities, particularly struggling with B-PER recognition ($0.65-0.70$ F1-score).

\noindent\textbf{LOCATION Entities:} All models perform consistently well on location entities, with CRF achieving the highest F1-scores ($0.84-0.85$ for B-LOC). The fact that results for location entities are clear on most models means that these entities appear to have characteristic features that are easy for different algorithms to recognise.

\noindent\textbf{ORGANIZATION Entities:} Organisation recognition shows moderate performance across all models. CRF achieves the best performance (F1-score of $0.82-0.83$) and BiLSTM tends to predict more organisation entities than might be correct (F1$=0.80-0.81$).

\noindent\textbf{DATE Entities:} Date recognition proves challenging for all models, with F1-scores ranging from $0.62-0.77$ for B-DATE tags. The relatively low performance may be attributed to diverse date formats and expressions in Kurdish Sorani text. CRF shows the most consistent performance across both data splits.

\noindent\textbf{MISCELLANEOUS Entities:} MISC entities present the greatest challenge, with I-MISC tags achieving the lowest F1-scores across all models ($0.53-0.65$). This difficulty is caused by the fact that miscellaneous entities have unlike features and include many different types of concepts.

\subsubsection{Performance Patterns Analysis}

The results reveal several important patterns. First, CRF consistently outperforms other models across all entity types, demonstrating its effectiveness for Kurdish Sorani NER. Second, I-tags (inside entity tokens) generally show lower performance than B-tags (beginning entity tokens), particularly for MISC and PER entities, indicating challenges in maintaining entity boundaries.

The Bidirectional Long Short-Term Memory (BiLSTM) model has strong recall and lower precision for many entity types, suggesting it sometimes predicts too many entities. The BiLSTM-CRF hybrid model improves precision compared to standalone BiLSTM while maintaining reasonable recall but does not achieve the performance levels of traditional CRF or SVM approaches.

\subsubsection{\textbf{Error Categories}}

Analysis of prediction errors reveals three main categories. Boundary detection errors happen when models correctly find that an entity is present but incorrectly find its boundaries. Entity type confusion happens particularly between ORG and LOC entities, where contextual disambiguation proves challenging. Out-of-vocabulary entity misclassification affects all models, with MISC entities being frequently missed or misclassified as O (outside) tags.

\subsection{\textbf{Cross-Validation Results}}
In order to accomplish very strong testing and overcome the possible biases of the single train-test split, we performed 10-fold cross-validation experiments. The importance of the methodology is more apparent in the context of low-resource languages in which the limited availability of data may place the model at a high risk of variance in the performance of the model on various data distributions.

\begin{table}[ht!]
\centering
\caption{Optimal Performance Across 10-Fold Cross-Validation}
\label{tab:optimal_cv_performance}
\begin{tabular}{lcccc}
\toprule
\textbf{Model} & \textbf{Best Fold} & \textbf{Precision} & \textbf{Recall} & \textbf{F1-score} \\
\midrule
CRF & 6 & 0.8670 & 0.7991 & 0.8317 \\
SVM & 10 & 0.8460 & 0.7874 & 0.8156 \\
BiLSTM & 6 & 0.7147 & 0.7982 & 0.7541 \\
BiLSTM-CRF & 9 & 0.7930 & 0.7379 & 0.7645 \\
\bottomrule
\end{tabular}
\end{table}

\begin{table}[ht!]
\centering
\caption{Cross-Validation Performance Metrics (Mean $\pm$ Standard Deviation)}
\label{tab:cv_performance_stats}
\begin{tabular}{lccc}
\toprule
\textbf{Model} & \textbf{Precision} & \textbf{Recall} & \textbf{F1-score} \\
\midrule
CRF & $0.8406 \pm 0.0170$ & $0.7976 \pm 0.0109$ & $0.8184 \pm 0.0090$ \\
SVM & $0.8329 \pm 0.0102$ & $0.7704 \pm 0.0104$ & $0.8004 \pm 0.0088$ \\
BiLSTM & $0.6699 \pm 0.0258$ & $0.7982 \pm 0.0090$ & $0.7282 \pm 0.0137$ \\
BiLSTM-CRF & $0.7779 \pm 0.0139$ & $0.7207 \pm 0.0145$ & $0.7481 \pm 0.0117$ \\
\bottomrule
\end{tabular}
\end{table}

Analysis of cross-validation provides some significant findings about the stability of the models and the extent to which they can generalise. The CRF model continues its excellence by having the best mean F1-score (0.8184 $\pm$ 0.0090), which shows its high performance and consistency over the folds. The large value of the standard deviation ($\sigma = 0.0090$) shows high-quality generalisation ability, which is imperative to be actually deployed to application areas when the distributions of data might differ.\\
Importantly, the SVM model has the lowest variance in all metrics ($\sigma_{F1} = 0.0088$), which indicates remarkable stability even though its average F1-score is slightly lower than that of CRF (0.8004). The property is especially useful where such performance guarantees are essential. There is also an interesting twist to the BiLSTM model, it has the largest mean recall (0.7982 $\pm$ 0.0090) but the largest degradation in precision (0.6699 $\pm$ 0.0258) so it ends up with the lowest overall F1-score. The high precision variance ($\sigma = 0.0258$) that implies sensitivity to the composition of the 
training data is typical to neural methods in low-resource scenarios.\\

The hybrid BiLSTM-CRF model shows mediocre performance (F1-score $= 0.7481 \pm 0.0117$), positively addressing some of the precision problems of the pure BiLSTM model, at the cost of a reasonable recall. Yet, it does not perform as well as classic machine learning methods, which is an indicator that the advantages of neural layouts are not so high in our low-resource setting, where the number of annotated tokens is $64,563$. The results reproduce recent low-resource NLP observations that traditional techniques can be superior to their deep learning counterparts in situations where there is a low amount of training data.
\section{Further discussion on results and their implications}
\label{sec:discussion}

We stress the fact that traditional machine learning algorithms in our comparative phase outperform deep learning models. Specifically, CRF achieves F1-scores of $0.8168$ and $0.8252$, surpassing BiLSTM, which records values of $0.7155$ and $0.7063$ on various data subsets. This pattern is particularly significant, aligns with recent research in low-resource NLP \cite{Li_2022} and warrants consideration both theoretically and practically.

\subsection{Theoretical Implications of Model Performance}
The superiority of CRF and SVM models over their corresponding neural counterparts can be justified by the underlying differences in how limited training data is handled by these types of architectures. Traditional models have characteristics resembling high inductive bias and low variance, enabling more stable learning in the presence of sparse annotations. The structured prediction capabilities of CRF include strong label transition constraints, which discourage the generation of incorrect BIO label sequences that often occur when using unconstrained neural models. This is an architectural advantage that is particularly beneficial in scenarios lacking sufficient training examples to learn these constraints implicitly.

The bias-variance tradeoff is observed explicitly in our findings. BiLSTM models achieved competitive recall scores ($0.7773–0.7982$) but significantly lower precision ($0.6521–0.6699$), attributed to overgeneralisation in settings with insufficient training examples. The number of $64,563$ tokens in \textbf{AgaCKNER} is too small to reach the critical dataset size at which neural architectures typically demonstrate their representational strengths, often estimated to be in the hundreds of thousands of annotated tokens for morphologically rich languages. This finding supports recent work suggesting that the sample complexity of neural models presents an inherent challenge in low-resource contexts.

Another explanatory factor is feature engineering. The advantage of traditional models is that they rely on manually designed features that incorporate linguistic expertise about the Kurdish Sorani language, such as morphological patterns, context windows, and sub-word properties. These characteristics can be considered a form of inductive bias that compensates for the scarcity of annotated data, whereas neural models must learn similar representations from scratch.

\subsection{Language-Specific Considerations}
Kurdish Sorani also presents specific challenges that enhance the relative benefits of traditional models. The morphological complexity of the language leads to a high out-of-vocabulary (OOV) rate, which has a disproportionately negative impact on neural models. The agglutinative morphology produces numerous surface forms that represent semantically identical items, requiring models to generalise across these variations. This form of variation is more effectively handled using explicit morphological features, as leveraged by CRF and SVM, than through the distributed representations used by BiLSTM.

The lack of consistent capitalisation in Kurdish Sorani removes one of the strongest signals for named entity recognition found in languages written with Latin scripts. This further suggests that conventional models benefit more from alternative feature sets, while neural models struggle without this key signal. Additionally, named entity borrowing from Arabic and English introduces code-switching phenomena, which are more reliably captured by rule-based or feature-engineere

\subsection{Entity-Level Performance Analysis}
The difference in performance among the entities discloses the systematicity of model behavior. \textbf{PERSON} entities received the best recognition rates across all models (F1-score: $0.65-0.92$), likely due to the common naming structures, honorific denominations, and patronymic tags prevalent in Kurdish culture. Overall, the high performance of the CRF model on \textbf{PERSON} entities (F1-score: $0.85-0.90$ on B-PER) establishes the effectiveness of sequential modeling in extracting multi-token person names.

\textbf{ORGANIZATION} and \textbf{LOCATION} entities were comparatively more complex and were often confused with one another. This confusion highlights an inherent ambiguity in Kurdish Sorani, where many organisations are named after locations and vice versa.\textbf{MISCELLANEOUS} entities consistently achieved lower recognition performance (F1-score: $0.53–0.65$ for I-MISC), indicating that this category is the most heterogeneous. The lack of specific linguistic markers and the broad range of concepts it covers contribute to this difficulty.

\textbf{DATE} entity recognition posed a moderate challenge (F1-score: $0.62-0.77$), largely due to the use of diverse calendar systems in Kurdish Sorani texts, including both local (municipal) and Western formats. The CRF model's ability to capture sequential dependencies was particularly advantageous in handling temporally complex expressions.

\subsection{Implications for Low-Resource NLP}
Our findings challenge the prevailing assumption that neural architectures are universally applicable and consistently superior in handling NLP tasks. In languages where the annotated volume is less than $100,000$ tokens, the cost-benefit ratios of traditional machine learning may be even more preferable, as they use fewer computational resources to achieve better performance. This produces far-reaching consequences for the democratisation of NLP technology in underserved language groups.

Small standard deviations in cross-validation (\( \sigma = 0.0088 \) for SVM) indicate the stability of traditional models, suggesting greater reliability in deployment scenarios. Such stability is essential for practical applications targeting Kurdish speakers, where stable performance is more valuable than marginal improvements in accuracy.

\section{Conclusions}\label{sec:conclusions}

We can conclude that the creation of \textbf{AgaCKNER} dataset makes a substantial contribution in the field of NLP and helped address one of the most important gaps in low-resources language through the specific case of Kurdish Sorani. 

Specifically for the Kurdish Sorani language, this work facilitates future research in information extraction, machine translation, and cross-lingual transfer to help grow the capacity of NLP for the Kurdish-speaking population.

At a more general level, the annotation methodology at the base of our system, which uses bespoke tools and ensures strict quality control, provides a replicable structure for the development of NER datasets, also for other low-resource languages, and the equal distribution of the five types of news adopted in our work contributes to domain diversity. The latter can be further magnified in future work by incorporating additional genres.

Furthemore, our analysis of data splits and cross-validation shows that traditional machine learning can be more effective and cost-efficient than neural models for languages with limited data. The CRF model's F1-score of 0.825 highlights its reliable performance for languages like Kurdish, where stability is more important than minor accuracy improvements. \textit{This is particularly important for democratising NLP technology for underserved language groups.} Also, we show that classical methods grounded in linguistic expertise remain relevant in the deep learning era.

Overall, the similarity in the performance patterns between the different configuration settings reinforces the generalisability of our findings beyond this particular dataset.

\subsection{Limitations and Future Directions}
 Creating AgaNER using only news articles may limit its applicability to areas like social media or historical records. While a sample size of $64,563$ tokens is a good starting point, it is small compared to high-resource corpora in other languages. Future research will consider semi-supervised and transfer-learning techniques to use larger Kurdish corpora.

Moreover, the effectiveness of baseline models suggests investigating hybrid architectures that combine feature engineering and neural elements. Using Kurdish-specific pre-trained embeddings or multilingual models with similar phonologies to Kurdish, could be another promising direction to explore to narrow the performance gap between traditional and neural models.

We also highlight a need for assessment methods that account for morphological differences, as traditional exact-match evaluations can unfairly penalise models for correctly tagging entities with slight morphological variations present in some languages.

This study has shown a result that the suboptimal conventional models such as CRF may still attain a higher performance with the Kurdish Sorani NER, but the BiLSTM-CRF based architecture has great potential in the future. Its learning of complex contextual patterns may be a powerful basis in more sophisticated neural networks and deeper linguistic analysis in more resource-richer situations.

Please refer to Journal-level guidance for any specific requirements.

\section*{Declarations}

\begin{itemize}
\item The dataset produced and used for this article is freely available from the repository \cite{agackner_mendeleydataset}. 
\item The code is freely available from the GitHub repository \cite{agackner_allinone}.  
\end{itemize}

\bibliography{biblio}

\end{document}